\documentclass[10pt,twocolumn,letterpaper]{article}

\usepackage{cvpr}
\usepackage{times}
\usepackage{epsfig}
\usepackage{graphicx}
\usepackage{amsmath}
\usepackage{amssymb}

\usepackage{bm}
\usepackage{tabularx}
\usepackage{booktabs}
\usepackage{enumitem}

\cvprfinalcopy 

\def\cvprPaperID{****} 

\newcommand{\labelA}{Soft Assigned Multi-Label}

\ifcvprfinal\fi
\begin{document}

\setlength{\textfloatsep}{5pt}

\title{Towards an Unequivocal Representation of Actions}

\author{Michael Wray\\
University of Bristol\\
\and
Davide Moltisanti\\
University of Bristol\\
{\tt\small firstname.surname@bristol.ac.uk}
\and
Dima Damen\\
University of Bristol\\
}

\maketitle


\begin{abstract}
   This work introduces verb-only representations for actions and interactions; the problem of describing similar motions (e.g. `open door', `open cupboard'), and distinguish differing ones (e.g. `open door' vs `open bottle') using verb-only labels. 
   Current approaches for action recognition neglect legitimate semantic ambiguities and class overlaps between verbs (Fig.~\ref{fig:intro}), relying on the objects to disambiguate interactions. 
We deviate from single-verb labels and introduce a mapping between observations and 
multiple verb labels -- in order to create an Unequivocal Representation of Actions.
The new representation benefits from increased vocabulary and a soft assignment to an enriched space of verb labels.
We learn these representations as multi-output regression, using a two-stream fusion CNN.
The proposed approach
outperforms conventional single-verb labels (also known as majority voting)
on three egocentric datasets for both recognition and retrieval.
\vspace*{-8pt}
\end{abstract}

\section{Introduction}
Consider a collection of verbs one uses to describe preparing morning coffee: \textit{open, pick, put, turn, scoop, pour, fill, stir, close, etc.} 
Verbs represent important information about how we can interact with the world, yet -- especially in the English language -- are usually given context in the form of object(s) for disambiguation. The motion that is used to push a door is different to that of pushing a button and, as such, door and button are used to differentiate between the two motions (\textit{i.e.}~`push-door' vs `push-button'). 
This was recently highlighted in~\cite{sigurdsson2017actions}, where increased confusion has been reported by human annotators when given a singular verb label, compared to verb-noun labels.
However, this leads to motions being tied towards objects when in fact the same motion could be applied to different objects, 
\textit{i.e.}~opening a cupboard is similar to opening a microwave or a fridge.

\begin{figure}
\centering
\includegraphics[width=\columnwidth]{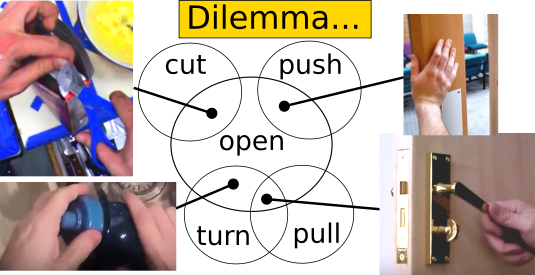}
\caption{Using single-verbs results in class overlaps.}
\label{fig:intro}
\end{figure} 

In this work we explore the idea of describing the action using a soft assignment over individually-ambiguous verb labels, yet keep it applicable for interactions with multiple objects. 
Take for example:
\textit{\{open, hold, turn, rotate\}}; by using multiple verbs, the motion is less ambiguous, yet is kept general to describe interactions with multiple objects, (\textit{e.g.}~jar, bottle, tap).
Note that we are not attempting to discover the objects being used; rather \textit{we seek a coherent representation of the action}, which can be used for recognition and retrieval tasks.
We focus on the egocentric domain, as object interactions are frequent and successive within a common environment.

\begin{figure*}[t]
  \begin{center}
  	\includegraphics[width=\textwidth]{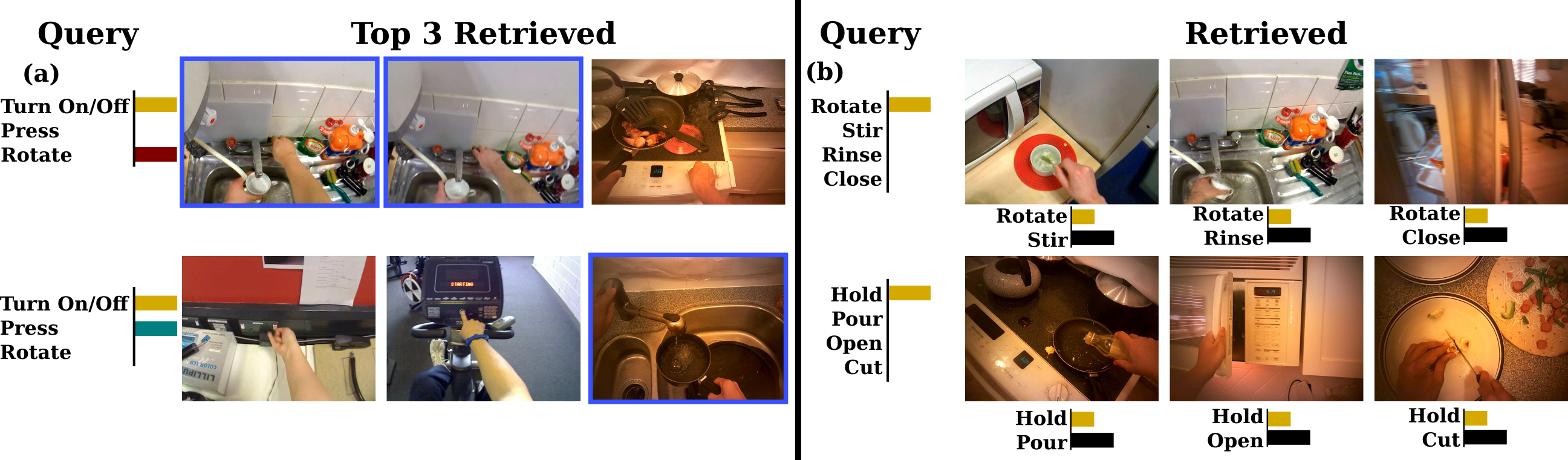}
  \end{center}
  \caption{Benefits of using multi-verb labels. (a) The labelling method is able to distinguish between turning on/off a tap by rotating and pressing (highlighted in blue). (b) Verbs such as rotate and hold can be learned via context from other actions.}
  \label{fig:main_example}
\end{figure*}

We propose benefits of using multiple verbs in Fig.~\ref{fig:main_example}.
In Fig.~\ref{fig:main_example}(a), we query our predicted representations using the verbs `turn-on/off', combined with one other verb (`rotate' vs `press').
The proposed unequivocal representation can make the distinction between a tap closed by rotating [first row in blue] and one by pressing [second row in blue], whereas neither single-verb nor verb-noun labels can.
 Models trained using the proposed representation can learn an enriched space of verb labels.
 In Fig.~\ref{fig:main_example}(b), different interactions can be retrieved using a common sub-action.

While we note that multi-label representations have become increasingly common for object recognition~\cite{wang2016cnn, wei2014cnn}, using multiple verbs to describe an action is under-explored for video understanding.
Previous datasets, egocentric~\cite{Fathi2012,wray2016sembed,de2008guide} and non-egocentric~\cite{kay2017kinetics,Kuehne11,sigurdsson2016hollywood,soomro2012ucf101} are annotated with a pre-selected number of verbs and commonly evaluated with classes defined as verb-noun pairs.
A few works attempt verb-only labels~\cite{Singh_2016_CVPR, wray2016sembed}, with both noting the difficulty and ambiguity of using single verb only labels.
Khamis and Davis~\cite{khamis2015walking} do use multi-verb labels in action recognition. However, they use a small amount of verbs (10) which describe non-overlapping actions, whereas we focus on using verbs which describe a single action.

We next present the proposed representation. 
It is crowdsourced, as in~\cite{chao2015benchmark,gella2016unsupervised,ronchi2015describing}, and evaluated using two-stream CNN~\cite{feichtenhofer2016convolutional}. We present results for classification and retrieval using three egocentric datasets.

\section{The Unequivocal Representation of Actions}
\label{sec:annotations}

\begin{figure}[t]
  \begin{center}
  	\includegraphics[width=\columnwidth]{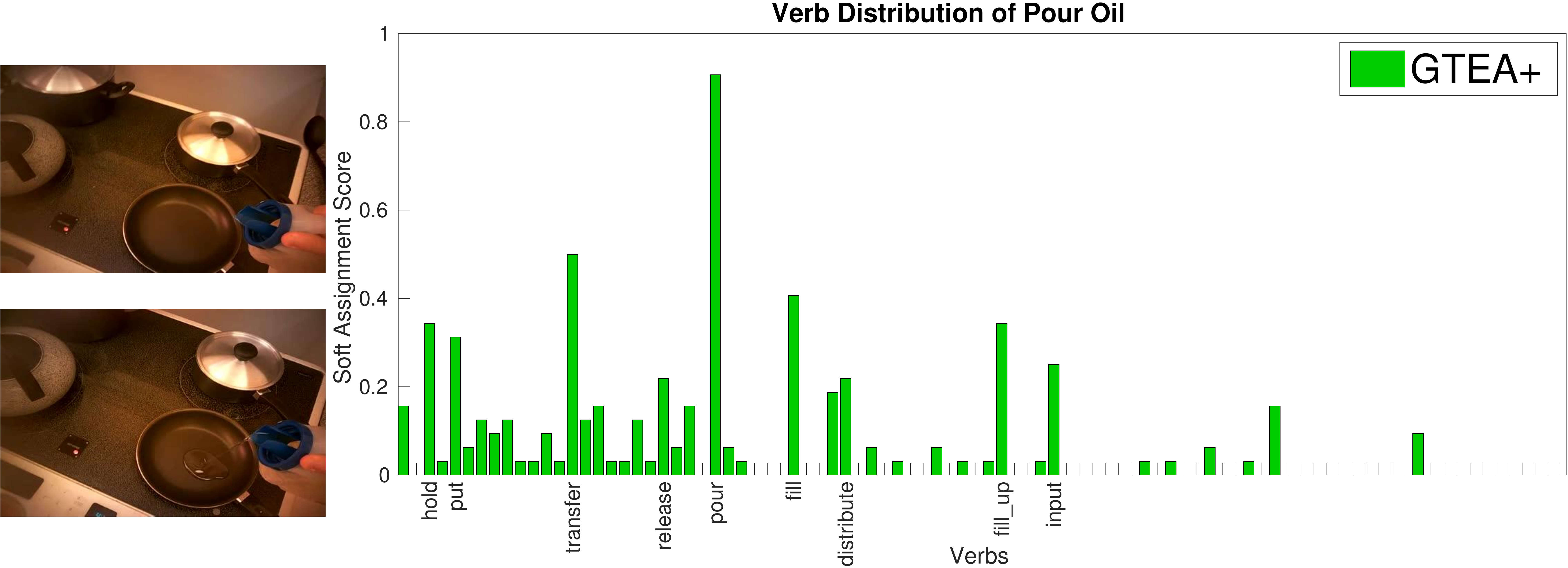}
  \end{center}
  \caption{Example annotation for the Pour Oil class from GTEA+.}
  \label{fig:annotation}
\end{figure}

In this section, we define the proposed representation that assigns multiple verb-only labels to action segments, in order to reduce single-verb ambiguity.
We use annotations as collected in~\cite{wray2017improving} for the three public datasets~\cite{wray2016sembed,de2008guide,Fathi2012}.
The annotations were collected per class with multiple annotators choosing which verbs, out of a list of 90, were applicable for the video (see Fig.~\ref{fig:annotation}).

\begin{figure*}[t]
  \begin{center}
     \includegraphics[width=\textwidth]{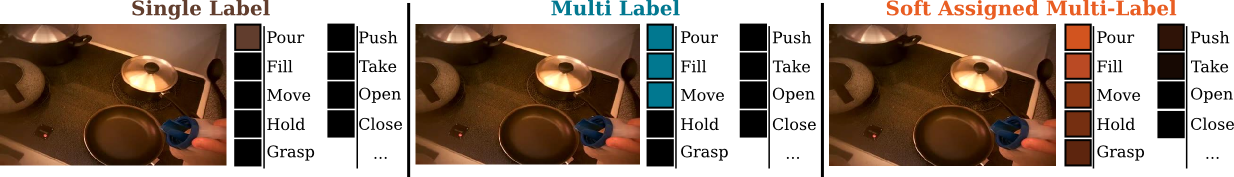}
  \end{center}
  \caption{
  In single verb labelling, only one verb can be chosen, often excluding valid labels. Multi-Label increases the vocabulary size and allows multiple verbs with equal importance to describe the same action. \labelA~increases the pool of verbs even further and uses  soft assignment for each. }
  \label{fig:motivation}
\end{figure*}


\noindent \textbf{Definitions:}\hspace{3pt} When using a \textbf{Single-verb Label (SL)}, each video $x_i \in X$ has a corresponding label $\bm{y_i} \in Y$ where $\bm{y_i}$ is a one-hot vector, over verbs $V=\langle v_j \rangle$. To minimise class overlaps, 
a small set of semantically distinct verbs are typically used. 
We use majority voting to create \textbf{SL}.

Alternatively, a \textbf{Multi-verb Label (ML)}, 
$\bm{y_i}=\langle y_{i,j} \in \{0,1\} \rangle$ is a binary vector over $V$.
Multiple verbs can be used to describe the video, e.g. `pour' and `fill'.
Hard assignment though can be problematic for sub-actions. Verbs such as `hold' do not fully describe the object interaction yet cannot be ignored as irrelevant. 
We construct \textbf{ML} as a binary vector where each verb above a threshold of $50\%$ is set to 1.

In introducing the \textbf{Soft Assigned Multi-Label~(SAML)}, we wish to increase the size of $V$ while accommodating sub-actions. 
Soft assignment offers a ranking of verb labels, ordered by the ability of the label, to sufficiently describe the ongoing interaction.
In the \labelA, each video $x_i$ will have a label vector 
$\bm{y_i}= \langle y_{i,j} \in [0,1] \rangle$ over $V$.
For two verbs $v_j, v_k$, in SAML, $y_{i,j} > y_{i,k} > 0$ when the first verb is more commonly used to describe the action in $x_i$, while $y_{i,k}$ is still a valid/relevant label.
We normalise the responses by the number of annotators to get the soft assignment score.

\noindent \textbf{Note:} Acquiring the proposed representations from semantics is potentially challenging.
Some verbs will be related semantically; e.g. `hold' and `grasp' are synonyms. Others are related via context, e.g. `pour' and `fill' are linked depending on the viewpoint (\textit{Is the bottle being poured?} or \textit{Is the cup being filled?}). Finally, we have sub-actions, e.g. the user must `hold' the bottle to be able to `pour' its contents. 
While these relationships can be explicitly stated, they are \textit{interestingly} not available in lexical databases 
and hard to discover from public corpora.
We study two commonly used sources of semantic information, {W}ord{N}et~\cite{miller1995wordnet} and {W}ord2{V}ec~\cite{mikolov2013efficient} embeddings, showing their limitations:

\vspace*{-1pt}
\resizebox{0.9\columnwidth}{!}{
  \begin{tabular}{ll|c|c}
    & &\textbf{\hspace{8pt}WordNet\hspace{8pt}} &\textbf{\hspace{8pt}Word2Vec\hspace{8pt}}\\ \hline
   synonyms &(e.g. `hold'-`grasp') &$\surd$ &$\times$ \\
   context-related &(e.g. `fill'-`pour')&$\times$ &$\times$\\
   sub-actions &(e.g. `hold'-`pour') &$\times$ &$\times$\\ 
  \end{tabular}}

\noindent \textbf{Learning:}
For each of the three labelling approaches (Fig~\ref{fig:motivation}), 
we wish to learn a function, ${\phi : \mathcal{W} \rightarrow \mathbb{R}^D}$ which maps a video representation $\mathcal{W}$ onto labels with ${D=|V|}$. For brevity we define $\bm{\hat{y_i}} = \phi(x_i)$, 
$\hat{y}_{i,j}$ as the predicted value for verb $v_j$ of video $x_i$ and $y_{i,j}$ as the corresponding ground truth.
Typically, the single label (SL) is learned using a cross entropy loss of the softmax scores.
To learn the multi-label (ML) we use a sigmoid binary cross entropy loss as commonly used in multi-label classification~\cite{nam2014large}.

In the \labelA~(SAML) representation, each element in $\bm{y_i}$ can take any value in the range $[0,1]$. 
We formulate this as a multi-task learning problem as defined in~\cite{Rai2012simultaneously}, solved as a multi-output regression 
without any independence assumptions. 
We again use the sigmoid binary cross entropy loss. 
We consciously avoid a ranking loss as it only learns a relative order and does not attempt to approximate the representation.

\noindent \textbf{Prediction and Evaluation:}
We can use $\phi = \{\phi_{SL}, \phi_{ML}, \phi_{SAML}\}$ to predict the labels for a previously unseen input $x_i$. We next present two ways to evaluate $\phi$ for predicting SL, ML and SAML.


We can evaluate $\phi$ in its ability to find `relevant' verbs. Given a threshold $\alpha$ for what verbs are deemed relevant 
$V_i^\alpha = \{v_j : y_{i,j} \geq \alpha, \; \forall v_j \in V \}$,
The top $k$ predicted verb labels would then be $\hat{V}_i^\alpha = \{\hat{y}_{i,j} : \hat{y}_{i,j} \in top_k(\bm{\hat{y_i}}) \land k=|V_i^\alpha|\}$.
The accuracy can now be calculated as a percentage of the overlap between the predicted and ground truth verbs:
\vspace{-5pt}
\begin{equation}
	A(\alpha|\phi) = \frac{1}{|X|}\sum_i \frac{|V_i^\alpha \cap \hat{V}_i^\alpha|}{|V_i^\alpha|}
    \label{eq:vtk}
\end{equation}
Note that $A(\alpha|\phi_{SL})$, for any $\alpha$, matches traditional classification accuracy, making this metric comparable to V-N.

Additionally, we can treat the $\phi$ as an embedding function such that $\bm{y_i}$ and $\hat{\bm{y_i}}$ represents the ground-truth and predicted embeddings of video $x_i$ respectively. 
Any verb $v_j$ can be located as a vertex in the label space which corresponds to the one-hot vector with $v_j$ set to 1 and the rest to 0, which we refer to as $\mathbf{v}_j$.
We can thus define video-to-text retrieval as the ranking of verbs from closest to furthest based on the $L_2$ distance ($||\hat{\bm{y_i}} - \mathbf{v_j}||$), which we can compare to the true ranking ($||\bm{y_i} - \mathbf{v_j}||$).
Similarly, we can define text-to-video retrieval from a given label $\mathbf{v}_j$ as the ranking of all embedded predictions $\hat{\bm{y_i}}$ using the same distance metric.
Importantly, we can construct more interesting text-to-video query vectors that involve multiple verbs to describe videos. Assume $\mathbf{u}^n_i$ is a binary vector with $n$ verbs being set to 1, and the rest to zero.
We accordingly perform text-to-video retrieval on these multi-verb queries and compare the various labelling methods.
In order to evaluate both text-to-video as well as video-to-text retrieval, we use mean Averaged Precision (mAP) over all queries as used in~\cite{philbin2007object}.


\section{Experiments and Results}
\label{sec:experiments}

We evaluate on the three annotated egocentric datasets:
BEOID~\cite{Damen2014a} ($732$ video segments), CMU~\cite{de2008guide} ($404$ video segments) and 
GTEA+~\cite{Fathi2012} ($1001$ video segments) using the annotations defined in Sec.~\ref{sec:annotations}. 
All three datasets include videos of daily activities indoors, 
recorded using a head mounted camera.
For each dataset, $5$ cross-fold validation was used where we equally distribute videos from each class. 

\noindent\textbf{Implementation Details:}
We trained for 100 epochs and tested using the two stream fusion CNN model from~\cite{feichtenhofer2016convolutional}
, pre-trained on UCF101~\cite{soomro2012ucf101}.
The number of nodes in the last layer equals the number of verbs $|V| = 90$. 

\begin{table}[t]
\centering
\begin{tabular}{lllll}
\toprule
      & SL & ML & SAML & V-N           \\ 
\midrule
BEOID & 78.1                    & \textbf{93.0}           & 87.8                      & \textbf{93.5} \\ 
CMU   & 59.2                    & 74.1                    & 73.5                      & \textbf{76.0} \\ 
GTEA+ & 59.2                    & \textbf{71.9}           & 67.8                      & 61.2          \\ 
\bottomrule
\end{tabular}
\caption{Action recognition accuracy (\%) results compared to verb-noun(V-N) classes using Eq.~\ref{eq:vtk}.}
\label{tab:class_results}
\end{table}

\begin{table*}[h]
\centering
\small
\begin{tabularx}{\textwidth}{lXXXXXXXXXXXXX}
\toprule
& & \multicolumn{3}{c}{BEOID}                    & & \multicolumn{3}{c}{CMU}                      & & \multicolumn{3}{c}{GTEA+}                    & \\ 
\cmidrule{3-5} \cmidrule{7-9} \cmidrule{11-13}
& & SL            & ML            & SAML          & & SL            & ML            & SAML          & & SL            & ML            & SAML          & \\ \midrule
Single-Verb only   & & \textbf{0.85} & 0.65          & 0.83          & & \textbf{0.71} & 0.55          & 0.68          & & \textbf{0.73} & 0.46          & 0.62          & \\ 
$\alpha \ge 0.5$ ranking   & & 0.46          & \textbf{0.95} & 0.89          & & 0.40          & \textbf{0.82} & 0.74          & & 0.42          & \textbf{0.79} & 0.76         &  \\ 
$\alpha \ge 0.3$ ranking & & 0.34          & 0.64          & \textbf{0.92} & & 0.36          & 0.68          & \textbf{0.81} & & 0.32          & 0.60          & \textbf{0.75} & \\ \midrule
Avg. mAP& & 0.55          & 0.75          & \textbf{0.88} & & 0.49          & 0.68          & \textbf{0.75} & & 0.49          & 0.62          & \textbf{0.71} & \\ 
\bottomrule
\end{tabularx}
\caption{Video-to-Text retrieval results on the three datasets using mAP. SAML performs consistently well over different ground truth.}
\label{tab:mAP_results}
\end{table*}


\noindent\textbf{Action Recognition Accuracy:} We motivate our work by stating that multi-verb labels allow removing the ambiguity of the performed action compared to single-verb labels. We accordingly evaluate all datasets using the original ground-truth verb-noun classes, and report results in Table~\ref{tab:class_results} as V-N.
We train V-N using the same loss function as SL, but use the dataset's verb-noun class labels.
We show that indeed adding verbs decreases the ambiguity and produces comparable results to verb-noun classes on BEOID (-0.5\%), slight drop on CMU (-2\%) and outperforms Verb-Noun classes on the largest of the three datasets, GTEA+ (+10\%). Moreover, the reported accuracies on ML and SAML are significantly higher than the ambiguous single-verb labels on all datasets, despite the increase in the number of verbs as potential output labels.
We note that $\phi_{ML}$ consistently achieves the highest accuracy compared to $\phi_{SL}$ and $\phi_{SAML}$ suggesting that it is more suited for recognition.



\begin{figure*}[t]
\begin{center}
   \includegraphics[width=0.9\textwidth]{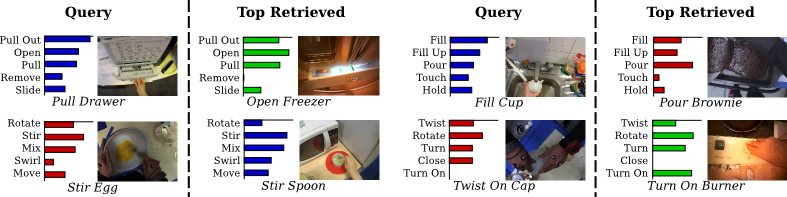}
\end{center}
   \caption{Examples of cross dataset retrieval of videos using either videos or text. Blue: BEOID, Red: CMU and Green: GTEA+.}
\label{fig:qual_vv_retrieval}
\end{figure*}
\begin{figure}[t]
  \begin{center}
  	\includegraphics[width=\linewidth]{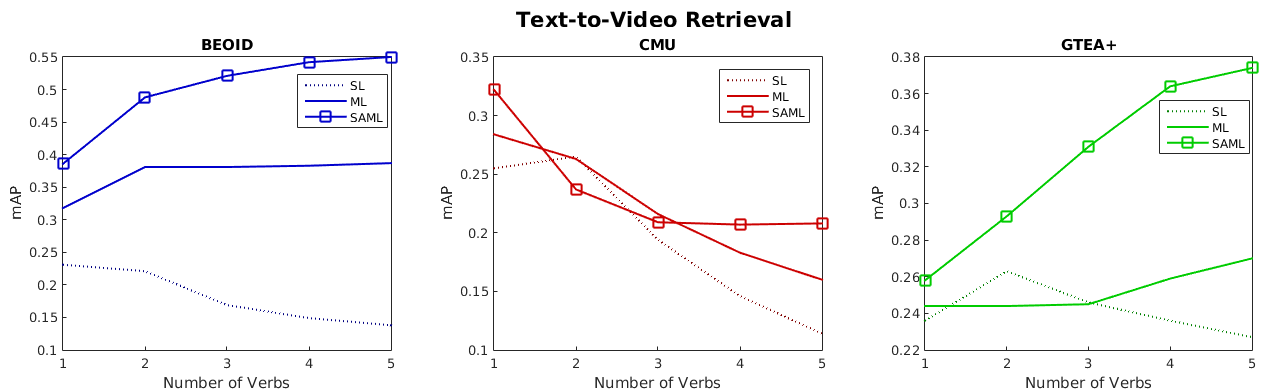}
  \end{center}
  \caption{Results of text-to-video retrieval across all three datasets using mAP and a varying number of verbs in the query.}
  \label{fig:retrieval_results}
\end{figure}

\noindent\textbf{Video-to-Text Retrieval Results:}
We next evaluate $\phi$ for video-to-text retrieval.
Table~\ref{tab:mAP_results} compares mAP of retrieval using each labelling method. 
$\phi_{SAML}$ has a solid performance consistently, even when compared to $SL$ in retrieving a single verb. Note that $\phi_{ML}$ has a good performance only at $\alpha \ge 0.5$ as this matches the labels on which it has been trained.
This is emphasised when we look at the average mAP scores of each $\phi$ across all datasets. SAML achieves the highest average mAP across all three datasets.

\noindent\textbf{Text-to-Video Retrieval Results:}
Figure~\ref{fig:retrieval_results} reports mAP results for text-to-video retrieval. In these results, we query the predicted representations using a binary vector $\mathbf{u}^n_i$ with $n$ verbs, and $n = \{1 \cdots 5\}$. Note that the verbs chosen are those that co-occur in the dataset to avoid antonyms like `open' with `close'. All possible combinations of $n$ co-occuring verbs have been tested.
As expected SL and ML perform best with either one or two verbs used as input with the mAP staying steady or dropping. SAML increases its mAP for BEOID and GTEA+ as the number of verbs increases -- outperforming SL and ML. This suggests that with the full vocabulary, the method is able to better learn the multi-verb representations for both the main verbs used to describe an action as well as any sub-actions.

\noindent\textbf{Cross-Dataset Retrieval:}
In this section we perform video-to-video retrieval, using SAML labels, across datasets. 
We first predict the representation of a video using $\phi_{SAML}$, then use this representation to find the top-retrieved video from a different dataset. For example, when querying using the video `stir egg' from CMU, the top-retrieved video is of `stir spoon' from BEOID. Interestingly, `pull drawer' from BEOID retrieves `open freezer' in GTEA+, as they both include the same motion, one being the drawer of a printer and the other the drawer of a freezer. In the last example, $\phi_{SAML}$ relates `twist-on cap' from CMU to `turn-on burner' from GTEA+ as both perform similar motions.

We earlier presented sample text-to-video cross-dataset retrievals in Fig.~\ref{fig:main_example}.
While we don't report cross-dataset quantitative results, we believe examples in Fig.~\ref{fig:qual_vv_retrieval} show the potential of the proposed representations beyond a single dataset.



\vspace*{-6pt}
\section{Conclusion and Future Work}
\vspace*{-2pt}

In this paper, we present the case for using multi-verb labels for action videos, and propose the Soft Assigned Multi-verb labels.
Compared to single verb-only labels, this offers an unambiguous representation of the interaction, embracing class overlaps. On the other hand, when compared to verb-noun labels, this representation generalises to multiple and unseen objects whilst still performing similarly comparably for recognition. 

The representations, learned using a two-stream fusion CNN, are able to predict the correct verb labels -- outperforming single-verb labels, for both recognition and retrieval. 
This representation can be useful of zero-shot or few-shot learning, predicting novel combination of verb labels. We will embark on assessing this next.

{\small
\bibliographystyle{ieee}
\bibliography{towards}
}

\end{document}